\documentclass[twocolumn,superscriptaddress,longbibliography]{revtex4-1}
\usepackage{color}
\usepackage[colorlinks=true,urlcolor=blue,citecolor=blue]{hyperref}
\usepackage{soul}
\usepackage{graphicx}
\usepackage{amsmath}
\usepackage{subfigure}
\usepackage{setspace}
\usepackage{hyperref}
\usepackage{url}
\usepackage[hyphenbreaks]{breakurl}

\sethlcolor{yellow}

\graphicspath{{figures/}}

\makeatletter
\newcommand*\bigcdot{\mathpalette\bigcdot@{.5}}
\newcommand*\bigcdot@[2]{\mathbin{\vcenter{\hbox{\scalebox{#2}{$\m@th#1\bullet$}}}}}
\makeatother

\begin{document}
	
	\title{Machine Learning by Unitary Tensor Network of Hierarchical Tree Structure}
	% \title{Machine Learning by Two-Dimensional Hierarchical Tensor Networks: A Quantum Information Theoretic Perspective on Deep Architectures}
	
	\author{Ding Liu}
	\affiliation{School of Computer Science and Technology, Tianjin Polytechnic University, Tianjin 300387, China}
	\affiliation{ICFO-Institut de Ciencies Fotoniques, The Barcelona Institute of Science and Technology, 08860 Castelldefels (Barcelona), Spain}
	\author{Shi-Ju Ran}
	\email{sjran@cnu.edu.cn}
	\affiliation{Department of Physics, Capital Normal University, Beijing 100048, China}
	\affiliation{ICFO-Institut de Ciencies Fotoniques, The Barcelona Institute of Science and Technology, 08860 Castelldefels (Barcelona), Spain}
	\author{Peter Wittek}
	\email{https://peterwittek.com/}
	\affiliation{University of Toronto, M5S 3E6 Toronto, Canada}
	\affiliation{Creative Destruction Lab, M5S 3E6 Toronto, Canada}
	\affiliation{Vector Institute for Artificial Intelligence, M5G 1M1 Toronto, Canada}
	\affiliation{Perimeter Institute for Theoretical Physics, N2L 2Y5 Waterloo, Canada}
	\author{Cheng Peng}
	\affiliation{School of Physical Sciences, University of Chinese Academy of Sciences, Beijing 100049, China}
	\author{Raul Bl\'azquez Garc\'ia}
	\affiliation{ICFO-Institut de Ciencies Fotoniques, The Barcelona Institute of Science and Technology, 08860 Castelldefels (Barcelona), Spain}
	\author{Gang Su}
	\affiliation{School of Physical Sciences, University of Chinese Academy of Sciences, Beijing 100049, China}
	\affiliation{Kavli Institute for Theoretical Sciences, and CAS Center of Excellence in Topological Quantum Computation, University of Chinese Academy of Sciences, Beijing 100190, China}
	\author{Maciej Lewenstein}
	\affiliation{ICFO-Institut de Ciencies Fotoniques, The Barcelona Institute of Science and Technology, 08860 Castelldefels (Barcelona), Spain}
	\affiliation{ICREA, Passeig Lluis Companys 23, 08010 Barcelona, Spain}
	
	\begin{abstract}
		The resemblance between the methods used in quantum-many body physics and in machine learning has drawn considerable attention. In particular, tensor networks (TNs) and deep learning architectures bear striking similarities to the extent that TNs can be used for machine learning. Previous results used one-dimensional TNs in image recognition, showing limited scalability and flexibilities. In this work, we train two-dimensional hierarchical TNs to solve image recognition problems, using a training algorithm derived from the multi-scale entanglement renormalization ansatz. This approach introduces mathematical connections among quantum many-body physics, quantum information theory, and machine learning. While keeping the TN unitary in the training phase, TN states are defined, which encode classes of images into quantum many-body states. We study the quantum features of the TN states, including quantum entanglement and fidelity. We find these quantities could be properties that characterize the image classes, as well as the machine learning tasks. 
	\end{abstract}
	
	\maketitle
	
	\emph{Introduction.}--- Over the past years, we have witnessed a booming progress in applying quantum theories and technologies to realistic problems including quantum simulators~\citep{TCZBDN12nphysRev} and quantum computers~\citep{S98QcompRev,E10QcompuRev,BAN11QcompuRev}. To some extent, the power of the ``Quantum" stems from the properties of quantum many-body systems. As one of the most powerful numerical tools for studying quantum many-body systems~\citep{VMC08MPSPEPSRev,O14TNSRev,O14TNadvRev,RTPCSL17TNrev}, tensor networks (TNs) have drawn more attention. For instance, TNs have been recently applied to solve machine learning problems such as dimensionality reduction~\citep{CLOPZ+17TNML1rev,CLOPZ+17TNML2rev} and handwriting recognition~\citep{stoudenmire2016supervised,han2017unsupervised}. Just as a TN allows the numerical treatment of difficult physical systems by providing layers of abstraction, deep learning achieved similar striking advances in automated feature extraction and pattern recognition using a hierarchical representation~\citep{lecun2015deep}. The resemblance between the two approaches is beyond superficial. At the theoretical level, there is a mapping between deep learning and the renormalization group~\citep{beny2013deep}, which in turn connects holography and deep learning~\citep{you2017machine,GS17MLholoarxiv}, and also allows to design networks from the perspective of quantum entanglement~\citep{LYCS17MLent}. In turn, neural networks can represent quantum states~\citep{carleo2016solving,CCXWX17TNML,HM17TNML,glasser2017neural}.

\begin{figure}[htbp]
		\begin{minipage}{\linewidth}
			\includegraphics[width=\textwidth]{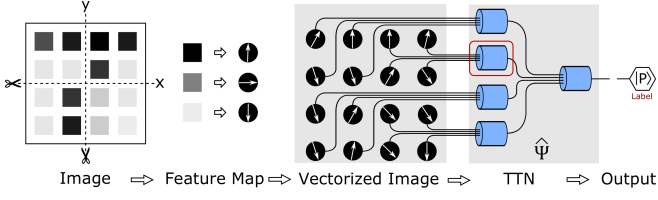}
		\end{minipage}
		%\hspace{1in}
		\caption{A $(4 \times 4)$ image of ``7'' is firstly vectorized to a product state by mapping each pixel to a $d$-dimensional vector, and then fed to the TTN denoted by $\hat{\Psi}$. The output is the vector after contracting with the TTN. The accuracy is obtained by comparing the output with the vectorized label $|p \rangle$. The environment tensor of, e.g., the tensor $T$ in the circle, is calculated by contracting everything after taking $T$ out. }
		\label{fig:TTN}
	\end{figure}

	In this work, we derive an quantum-inspired learning algorithm based on the multi-scale entanglement renormalization ansatz (MERA) approach~\citep{V07EntRenor,V08MERA,CDR08MERA,EV09MERAalgo} and hierarchical representation that is known as the tree TN (TTN)~\citep{Shi206TreeSimulation,MVSNL15TTN}. As shown in Fig.~\ref{fig:TTN}, the idea is first transform images to vectors living in a $d^N$-dimensional Hilbert space~\citep{stoudenmire2016supervised}, and then map the vectors (or ``vectorized images'') through a TTN (denoted by $\hat{\Psi}$) to predict the classes as outputs. Here $N$ is the number of pixels, $d$ is the dimension of the vector mapped from one pixel, and $D$ is the number of classes. A TTN of hierarchical structure \cite{cohen2016expressive} suits the two-dimensional (2D) nature of images more than those based on a one-dimensional (1D) TN, e.g., matrix product state~\citep{stoudenmire2016supervised,han2017unsupervised,novikov2016exponential}. Secondly, we explicitly connects machine learning to quantum quantities, such as fidelity and entanglement. Additionally, our algorithm allows to use unitary gates to construct the TTN, which makes it possible in the future to use quantum simulations/computations~\citep{vidal2003efficient, ZHXH+17MPSexp, Huggins2018Towards} to implement our proposal.
	
	The algorithm is tested on both the MNIST (handwriting recognition) and CIFAR (recognition of images) databases. The quantum many-body states that encode different classes of images can be defined from $\hat{\Psi}$ [see $|\psi_p\rangle$ in Eq. (\ref{eq-psip})], which is also in a TTN form. This is akin to the duality between probabilistic graphical models and TNs~\citep{robeva2017duality}. Combining with a common low-dimensional embedding method called t-SNE~\citep{van2008visualizing}, we find that the level of abstraction changes in a similar way as in a deep convolutional neural network~\citep{krizhevsky2012imagenet}, or a deep belief network~\citep{hinton2006fast}. The highest level of the hierarchy allows a clear separation of the classes. Finally, we study the fidelity and entanglement of the trained states. Comparing with the accuracy of the two-class classifiers, our results imply that the fidelity exhibits the difficulty of classifying the given two classes. As to the entanglement that is known to characterize the complexity of the TN schemes, moderate entanglement entropies [$S \sim O(1)$] are observed, implying that the TTN can efficiently represent the classifiers as slightly-entangled states.
	
	%\subsection*{Feature map}
	\emph{Tree tensor network and the algorithm.}--- Our approach begins from mapping each pixel $x$ to a $d$-component vector $\mathbf{v}(x)$. This is known as the feature map \citep{stoudenmire2016supervised} that is defined as
	\begin{eqnarray}
	v_s(x)=\sqrt{\binom{d-1}{s-1}}(\cos(\frac{\pi }{4}x))^{d-{s}}(\sin(\frac{\pi }{4}x))^{s-1}
	\label{eq-fmap}
	\end{eqnarray}
	where $s$ runs from $1$ to $d$. By using a larger $d$, the TTN has the potential to approximate a richer class of functions. With such nonlinear feature map, we project a gray-scale image (labeled as the $n$-th instance) from a scalar space to a $d^L$-dimensional vector space, where the image is represented as a product state of $L$ local $d$-dimensional vectors $|v^{[n]}\rangle = \prod_{j=1}^L |v^{[n,j]}\rangle$ (note $|v^{[n,j]}\rangle = \sum_s v^{[n,j]}_s(x) |s \rangle$ is the vector from the corresponding pixel [Eq. (\ref{eq-fmap})], $|s \rangle$ denotes the local basis of one spin, and $L$ is the number of pixels in the image.
	
	%\subsection*{MERA-inspired training algorithm}
	\label{sec-3}
	Let us introduce $\hat{\Psi}$ as a hierarchical structure of $K$ layers TN (see Fig.~\ref{fig:TTN}), whose coefficients are given by
	\begin{equation}
	% \hat{\Psi}_{\alpha_{1,1}\cdots\alpha_{L_K,p_4}} = \sum_{\{\alpha\}} \prod_{k=1}^{K} \prod_{p=1}^{L_k} T^{[k,p]}_{ \alpha_{k+1,p'} \alpha_{k,p_1} \alpha_{k,p_2} \alpha_{k,p_3} \alpha_{k,p_4}},
	\hat{\Psi}_{s_{1,1}\cdots s_{L_K,p_4}} = \sum_{\{s\}} \prod_{k=1}^{K} \prod_{m=1}^{L_k} T^{[k,m]}_{s_{k+1,m'} s_{k,m_1} s_{k,m_2} s_{k,m_3} s_{k,m_4}},
	\label{eq-TTN0}
	\end{equation}
	where $L_k$ is the number of tensors in the $k$-th layer and $K$ is the number of layers. The indexes contracted with $\{|v^{[n,j]}\rangle\}$ are called \textit{input bonds}, the top index corresponding to the label is called \textit{output bond}, and those in between are called \textit{virtual bonds}. Without losing generality, we assume all tensors are real.
	
	The output is a $D$-dimensional vector obtained by contracting $|v^{[n]}\rangle$ with the TTN as
	\begin{eqnarray}
	|\tilde{p}^{[n]}\rangle=\hat{\Psi}^{\dagger}| v^{[n]}\rangle
	\label{eq-Lout}
	\end{eqnarray}
	where $|\tilde{p}^{[n]}\rangle$ acts as the predicted label corresponding to the $n$-th sample. Based on these, we derive a highly efficient training algorithm inspired by MERA \footnote{The code of the implementation is available at \href{https://github.com/dingliu0305/Tree-Tensor-Networks-in-Machine-Learning}\\{https://github.com/dingliu0305/Tree-Tensor-Networks-in-Machine-Learning}}. The cost function to be minimized is chosen as
	\begin{eqnarray}
	f =-\sum_{n=1}^N \langle \tilde{p}^{[n]}| p^{[n]}\rangle.
	\label{eq-fcost0}
	\end{eqnarray}
	with $N$ the total number of samples. It can be interpreted as the summation of the inner product between $|v^{[n]}\rangle$ and $\hat{\Psi}|p^{[n]}\rangle$, which allows us to derive the MERA algorithm. In fact, Eq. (\ref{eq-fcost0}) can be deducted from the cost function of square error
	\begin{equation}
	f = \sum_{n=1}^N (|p^{[n]}\rangle-|\tilde{p}^{[n]}\rangle )^2,
	\label{eq-costf}
	\end{equation}
	Transform Eq. (\ref{eq-costf}) in the following form
	\begin{eqnarray}
	f= \sum_{n=1}^N (\langle v^{[n]}|\hat{\Psi} \hat{\Psi}^{\dagger}  |v^{[n]}\rangle -2 \langle v^{[n]}| \hat{\Psi} |p^{[n]}\rangle + 1).
	\end{eqnarray}
	The third term comes from the normalization of $|p^{[j]}\rangle$, and we assume the second term is always real. The dominant cost comes from the first term. We borrow the idea from the MERA approach to reduce this cost by imposing $\hat{\Psi}$ to be unitary, i.e., $\hat{\Psi}^{\dagger} \hat{\Psi} = I$. Then $\hat{\Psi}$ is optimized with $\hat{\Psi} \hat{\Psi}^{\dagger} \simeq I$ satisfied in the relevant subspace. In other words, we do not require an identity from $\hat{\Psi} \hat{\Psi}^{\dagger}$, but mean $\sum_{n=1}^N \langle v^{[n]}| \hat{\Psi} \hat{\Psi}^{\dagger}  |v^{[n]}\rangle \simeq \sum_{n=1}^N \langle v^{[n]}| v^{[n]}\rangle = N$ under the training samples.

	In MERA, a stronger constraint is used: all tensors in the TTN are required to be unitary (called isometries), satisfying $T^{\dagger}T = I$ (i.e., obtaining an identity by contracting the four down indexes of $T$ with its conjugate). With this constraint, the TTN gives a unitary transformation that satisfies $\hat{\Psi}^{\dagger} \hat{\Psi} = I$; it compresses a $d^N$-dimensional space to a $D$-dimensional one. This way, the first term becomes a constant, and we only need to deal with the second term. The cost function becomes
	\begin{eqnarray}
	f =- \langle v^{[n]}| \hat{\Psi} |p^{[n]}\rangle.
	\label{eq-fcost1}
	\end{eqnarray}
	Each term in $f$ is simply the contraction of the TN.
	
	The tensors in the TTN are updated alternatively to minimize Eq.~(\ref{eq-fcost0}). To update $T^{[k,m]}$ for instance, we assume other tensors are fixed and define the \textit{environment tensor} $E^{[k,m]}$, which is calculated by contracting everything in Eq.~(\ref{eq-fcost0}) after taking out $T^{[k,m]}$ (Fig.~\ref{fig:TTN})~\citep{EV09MERAalgo}. Then Eq.~(\ref{eq-fcost0}) becomes $f=-\text{Tr} (T^{[k,m]} E^{[k,m]})$. Under the unitary constraint, the optimal point is reached by taking $T^{[k,m]} = V U^{\dagger}$, with $V$ and $U$ calculated from the singular value decomposition (SVD) $E^{[k,m]} = U \Lambda V^{\dagger}$. Then we have $f = -\sum_{a} \Lambda_{a}$. The update of one tensor becomes the calculation of the environment tensor and its SVD. The scaling of the computational complexity is $O((\chi^{5}+d^{4}\chi)LN) $, with $\chi$ the dimension of virtual bonds, $d$ the dimension of input bonds, $L$ the number of pixels of one image, and $N$ the number of training samples.
	
	%\subsection*{Multi-class classification}
	The strategy for building a multi-class classifier is flexible. We here choose the ``one-against-all'' scheme. For each class, we label the training samples as ``yes'' or ``no'' and train one TTN so that it recognizes whether an image belongs to this class or not. The output [Eq.~(\ref{eq-Lout})] is a two-dimensional vector. We fix the label for a \textit{yes} answer as $|\text{yes}\rangle = [1,0]$. For the $p$ image classes, we accordingly have $p$ TTNs $\{\hat{\Psi}^{[p]}\}$ and define a set of TTN states as
	\begin{eqnarray}
	| \psi_p \rangle = \hat{\Psi}^{[p]}|\text{yes}\rangle.
	\label{eq-psip}
	\end{eqnarray}
	To recognize a given sample (e.g., $v^{[n]}$), we introduce a $p$-dimensional vector $\mathbf{F}^{[n]}$ with $F^{[n]}_{p}=|\langle v^{[n]}|\psi_p\rangle|$ that is the fidelity between $| \psi_p \rangle$ and the vectorized image to be classified. The position of its maximal element gives which class the image belongs to.
	
\begin{figure}[htbp]
		\centering
		\subfigure{
			\label{fig:cifar:a}
			\includegraphics[width=1.6in,height=1.4in]{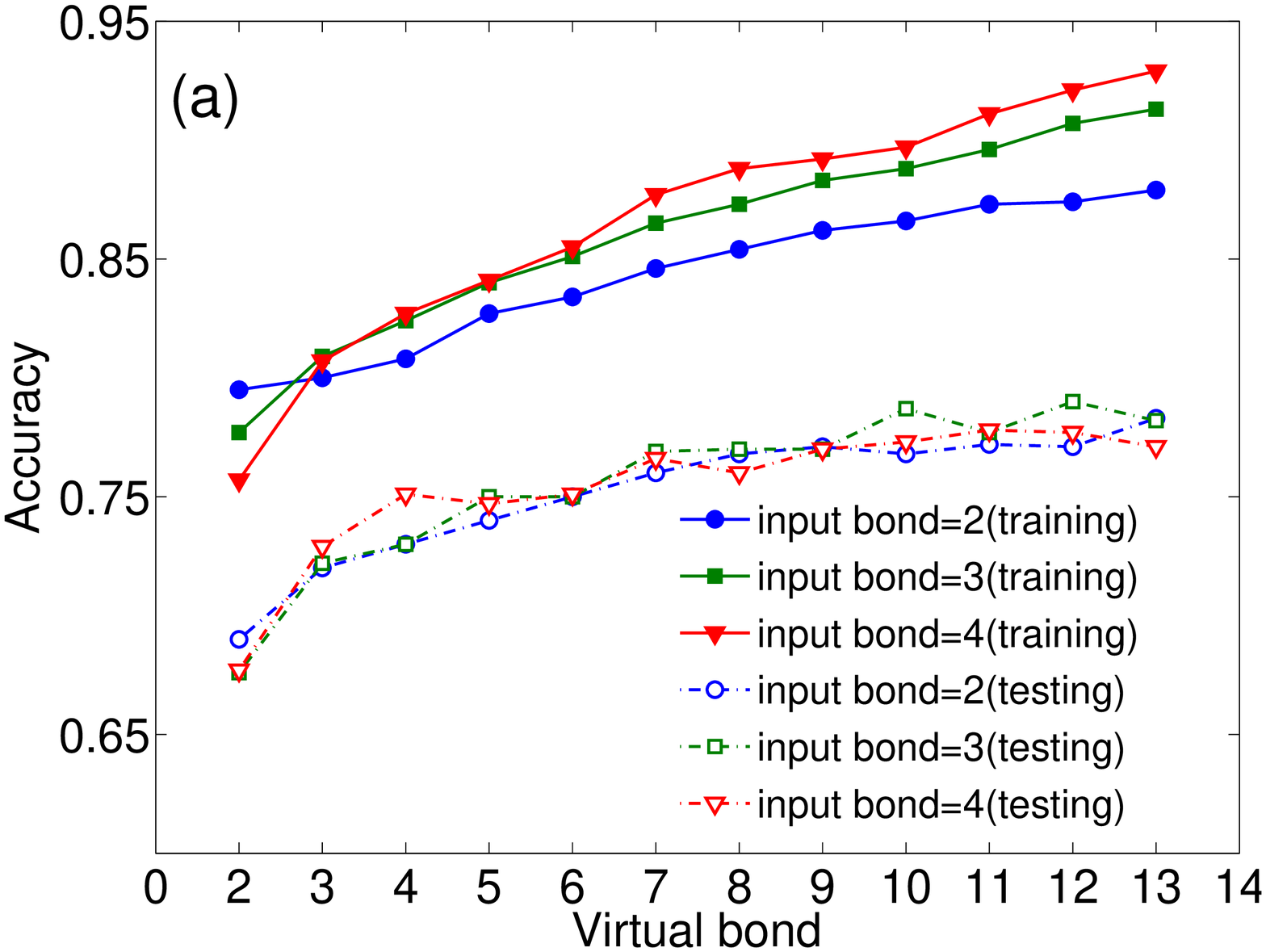}}
		\subfigure{
			\label{fig:cifar:b}
			\includegraphics[width=1.6in,height=1.4in]{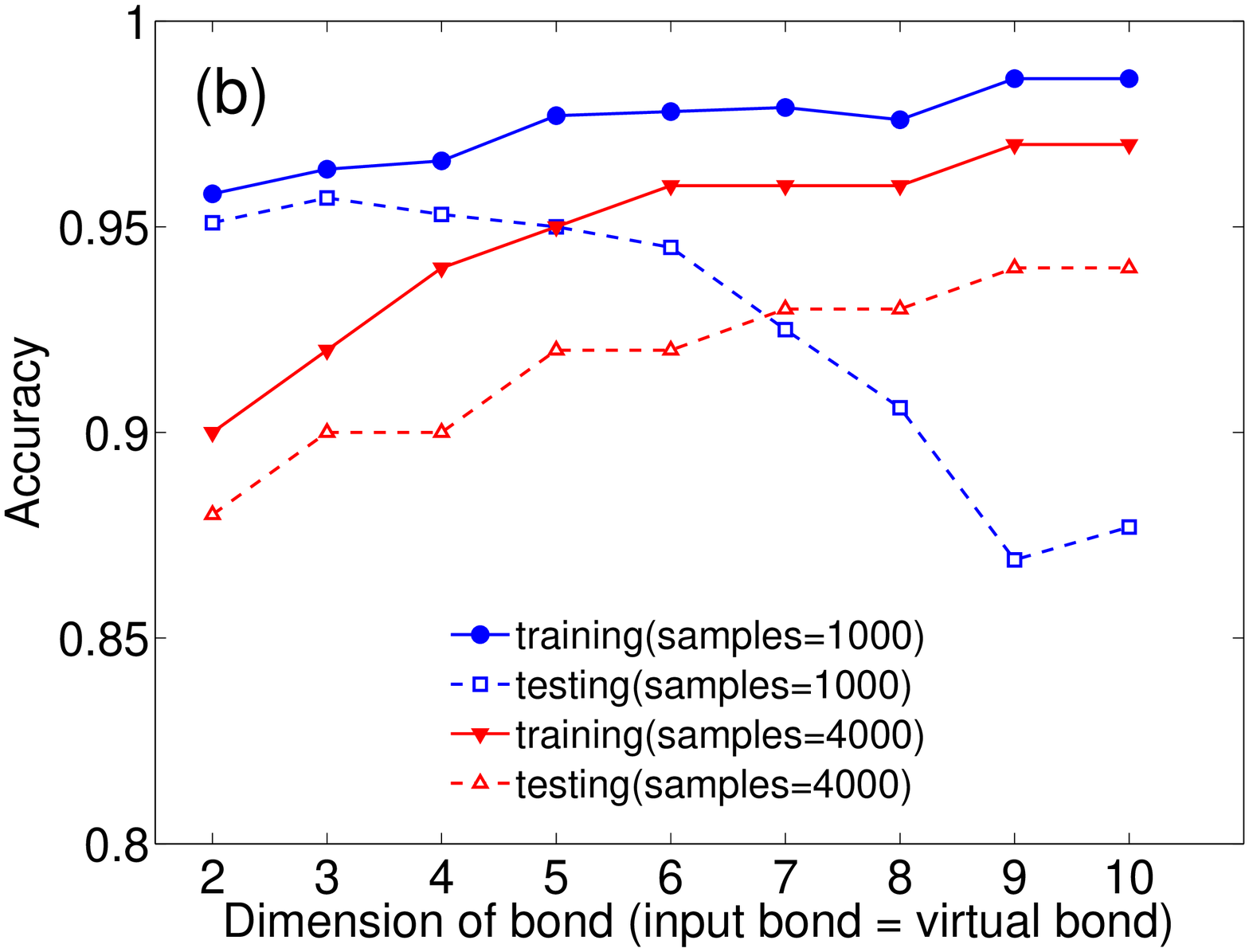}}
		\caption{(a) Binary accuracy on CIFAR-10 (Horses vs Planes) with 1000 training samples for each classes; (b) Training and test accuracy as the function of the bond dimensions on the MNIST dataset. The virtual bond dimensions are set equal to input bond dimensions. The number of training samples is taken 1000 or 4000 for each classes.}
		\label{fig:cifar}
	\end{figure}

	In our scheme, $|\psi_p \rangle$ can be understood as the many-body quantum state in the form of a TTN that approximates $| \tilde{\psi}_p \rangle = \sum_{n \in \text{p-th class}} |v^{[p]} \rangle$ (up to a normalization factor). In previous works, TTN form has been utilized in several inspiring but different ways in machine learning. For example, TTN is used to efficiently approximate a given higher-order tensor through tensor decompositions \cite{hackbusch2009new,grasedyck2010hierarchical}; by using delta tensors, one-to-one structural equivalence between TTN and convolutional arithmetic circuits was proposed, allowing for network design by entanglement \cite{LYCS17MLent}. Here, we obtain the TTN not by decomposing but by training, i.e., by minimizing the cost function. This is essentially different from the existing works: (1) our aim is to perform classification, instead of approximating a given tensor; (2) the tensors in the TTN are not restricted to a special form (like delta tensors) but are updated by the proposed MERA algorithm. Even though we have $|\psi_p \rangle = | \tilde{\psi}_p \rangle$ when the virtual bond dimensions of the TTN are sufficiently large, it will not give the optimal classification due to over-fitting [see Fig. \ref{fig:cifar:b} below].
	
	%\section*{Results}
	
	%\subsection*{Power of representation and generalization}
	\emph{Representation and generalization.}--- Representation power denotes here the parameter space a TN can actually reach. It refers to what kinds of functions can be realized by a TN. To demonstrate the representation power of the TTN, we use the CIFAR-10 dataset~\citep{krizhevsky2009learning}, which consists of 10 classes with 50,000 RGB images in the training dataset and 10,000 images in the testing dataset. Each RGB image was originally $32\times 32$ pixels. We transform them into gray-scale to reduce the algorithmic complexity without losing generality.
	
	Fig.~\ref{fig:cifar} (a) exhibits the relation between the representation power and the bond dimensions of the TTN $\hat{\Psi}$. Note that the representation power is reflected by the training accuracy (the accuracy on training dataset). Considering $\hat{\Psi}$ as a map that  projects the vectorized images from the $d^N$-dimensional space to the $D$-dimensional one, the limitation of the representation power of $\hat{\Psi}$ depends on the input dimension $d^N$ of $\hat{\Psi}$. Meanwhile, $\hat{\Psi}$ can be considered as an approximation of such an exponentially large map, by writing it into the contraction of small tensors. The dimensions of the virtual bonds determine how close $\hat{\Psi}$ can reach the limit.
	
\begin{figure}[htbp]
		\centering
		\subfigure{
			\label{fig:tsne:a}
			\includegraphics[width=1.6in,height=1in]{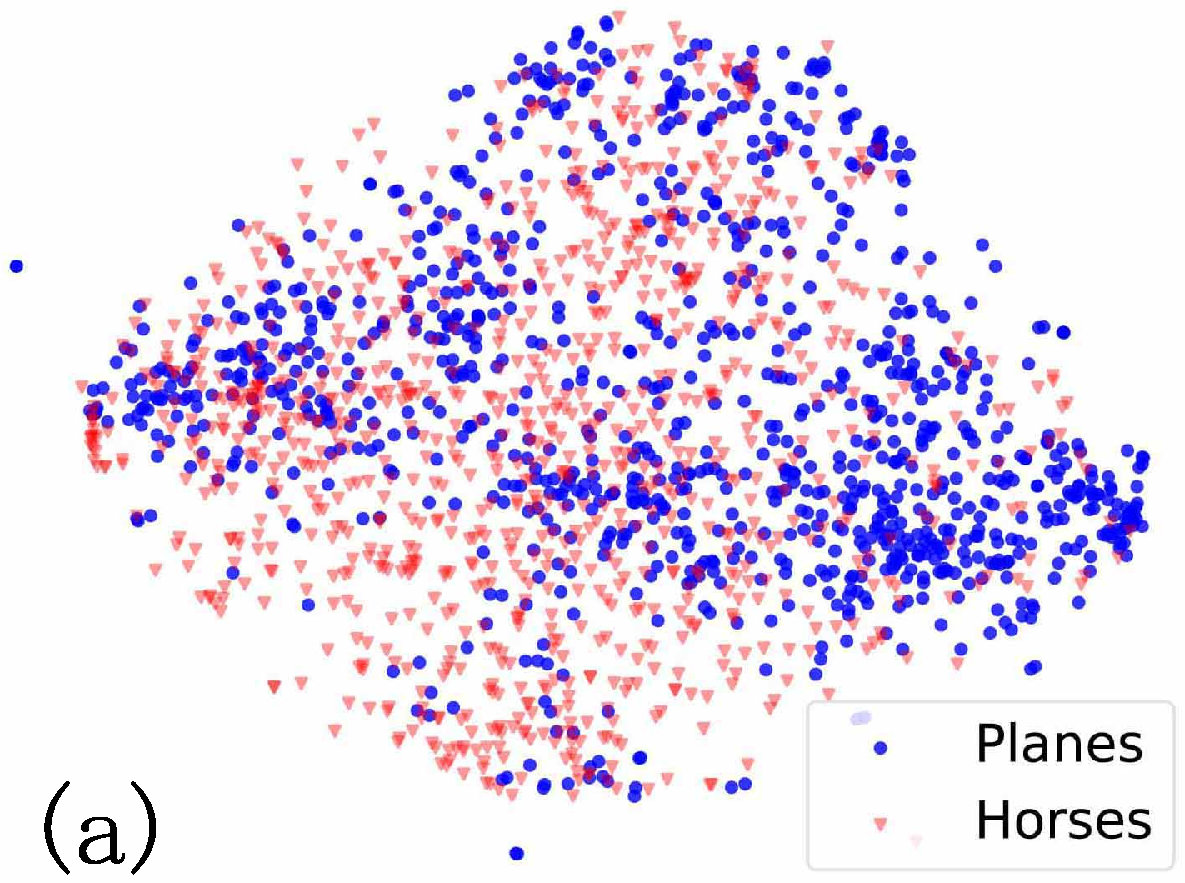}}
		\subfigure{
			\label{fig:tsne:b}
			\includegraphics[width=1.6in,height=1in]{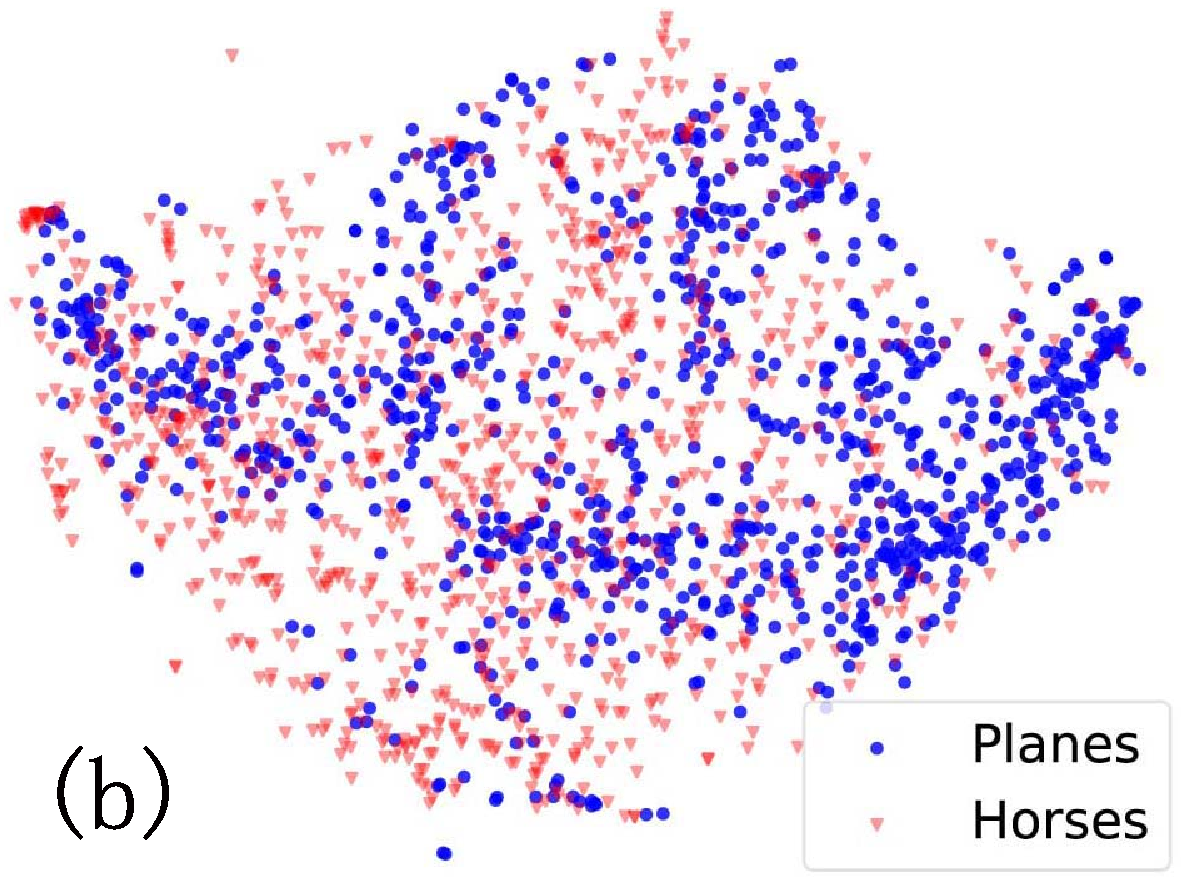}}
		\subfigure{
			\label{fig:tsne:c}
			\includegraphics[width=1.6in,height=1in]{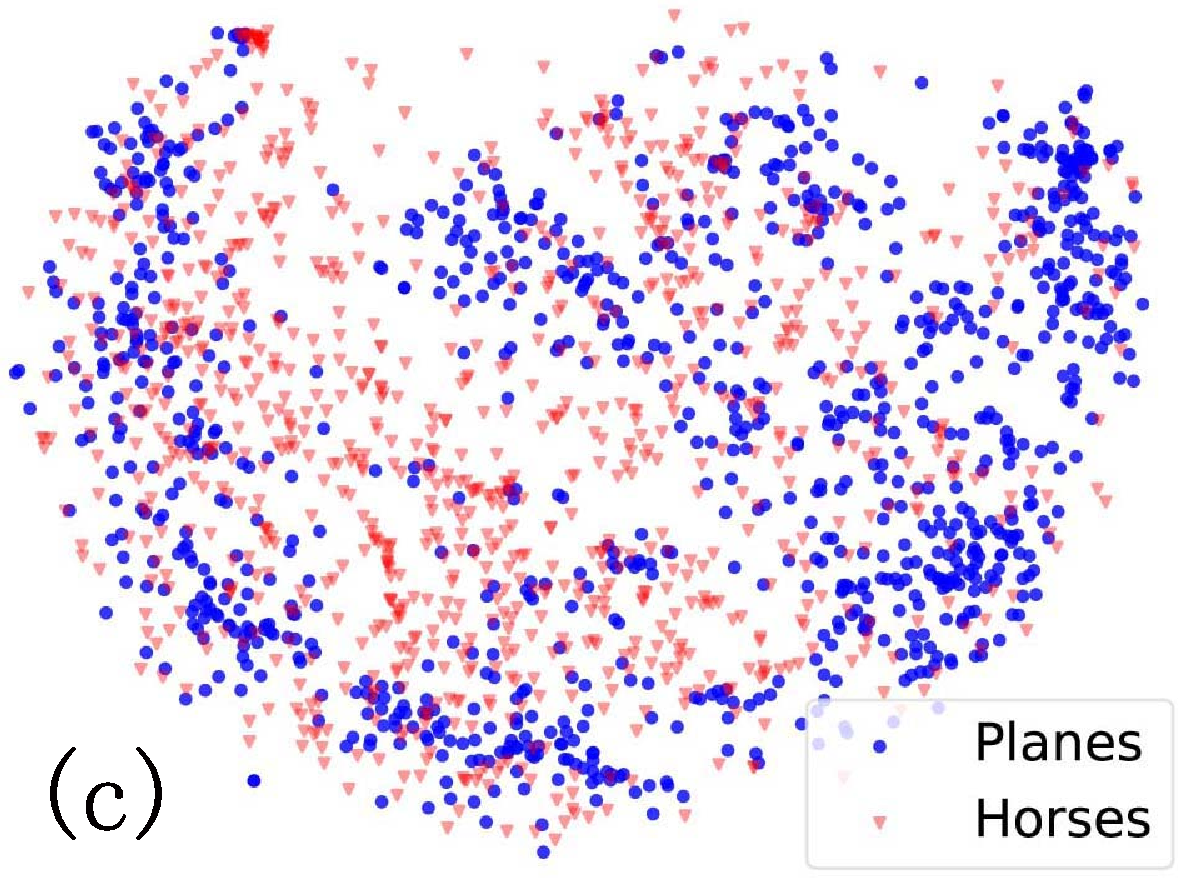}}
		\subfigure{
			\label{fig:tsne:d}
			\includegraphics[width=1.6in,height=1in]{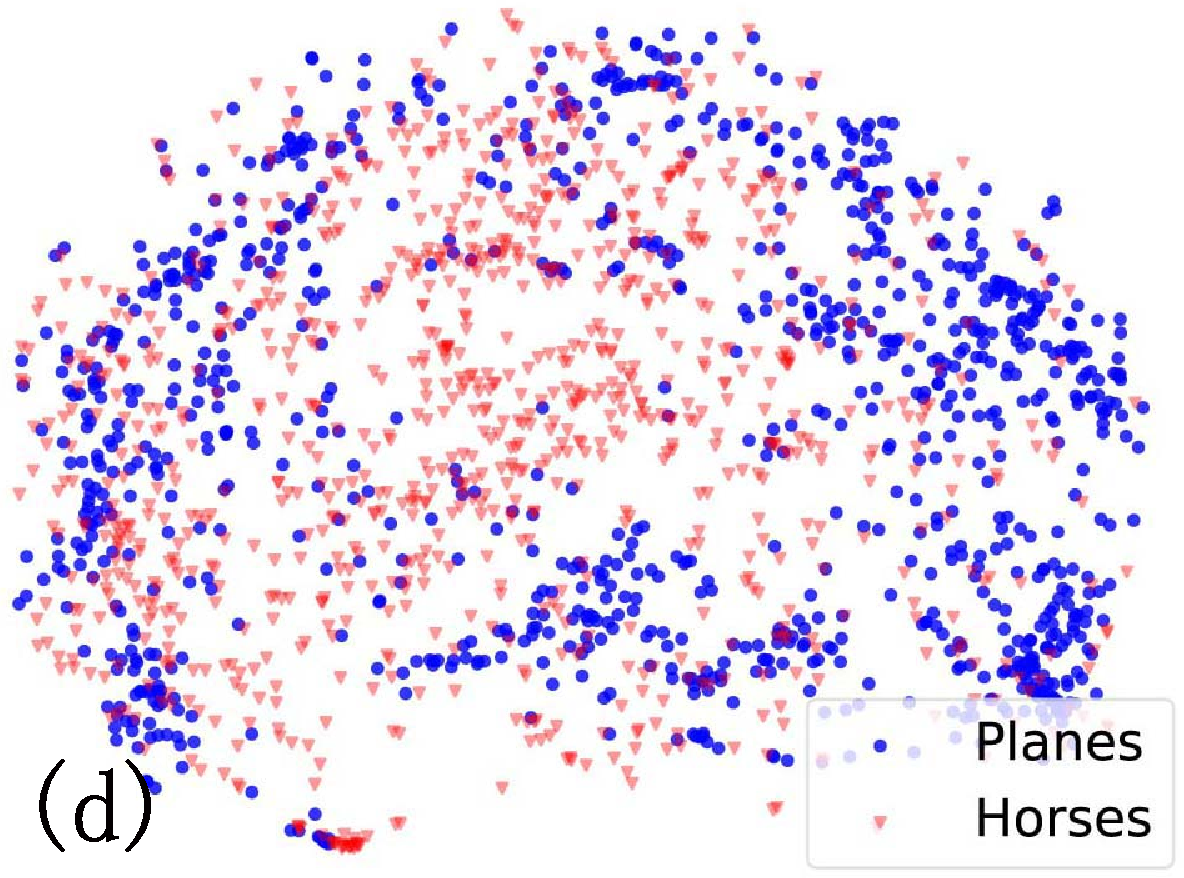}}
		\subfigure{
			\label{fig:tsne:e}
			\includegraphics[width=1.6in,height=1in]{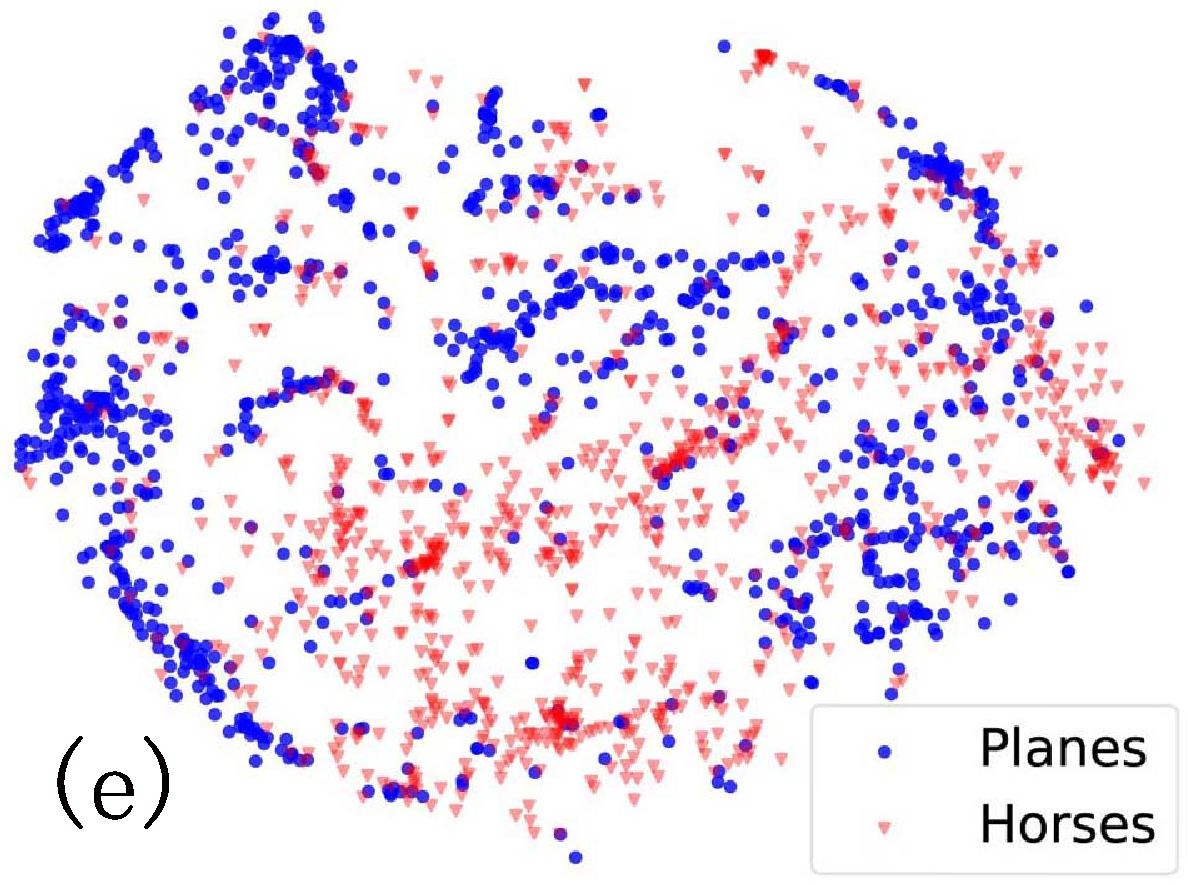}}
		\subfigure{
			\label{fig:tsne:f}
			\includegraphics[width=1.6in,height=1in]{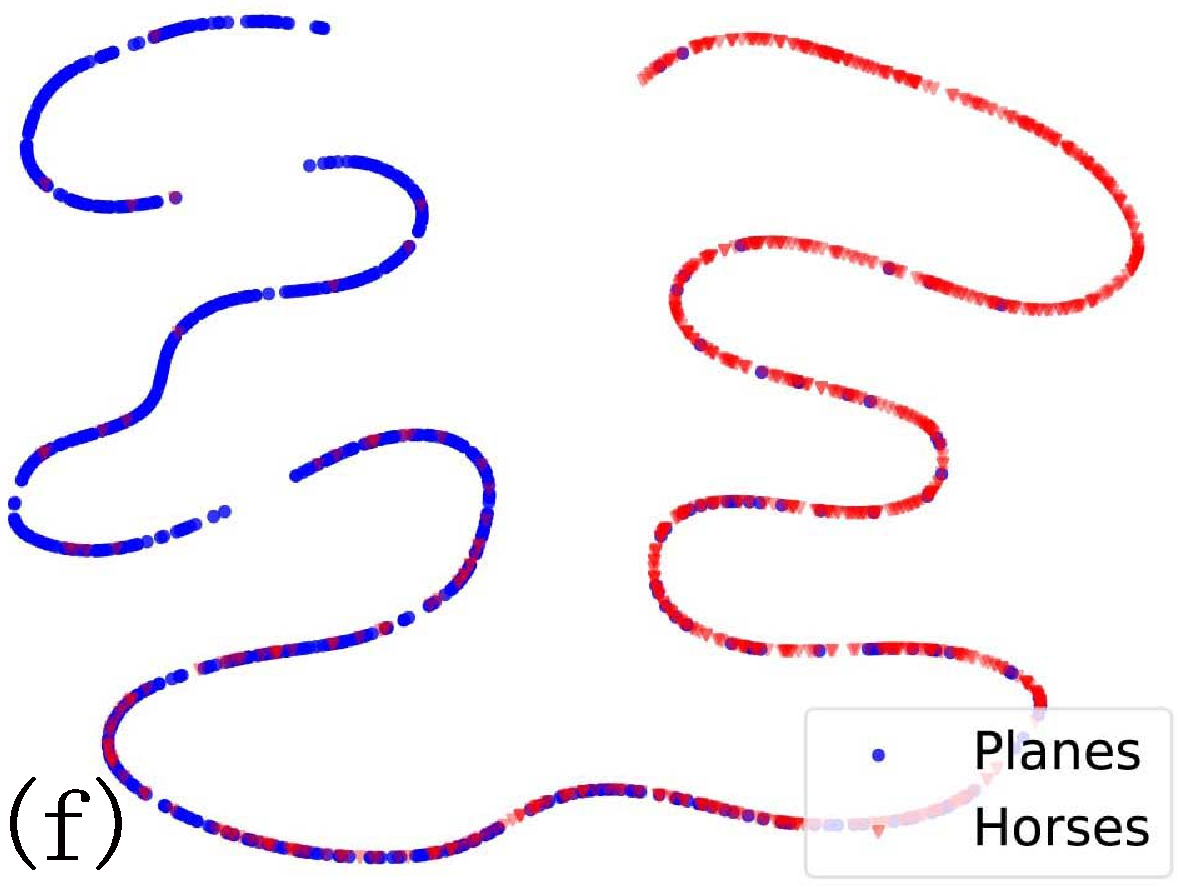}}
		\caption{Embedding of 1000 data instances for each classes of CIFAR-10 by t-SNE corresponding to each layer in the TTN: (a) original data distribution and (b) the 1st, (c) 2nd, (d) 3rd, (e) 4th, and (f) 5th layer.}
		\label{fig:tsne}
	\end{figure}

	The sequence of convolutional and pooling layers in the feature extraction part of a deep learning network is known to arrive at higher and higher levels of abstraction that helps separate the classes in a discriminative learner~\citep{lecun2015deep}. This is often visualized by embedding the representation in two dimensions using t-distributed stochastic neighbor embedding (t-SNE), which is a method for dimensionality reduction that is well suited for the visualization of high-dimensional datasets \citep{van2008visualizing}. If the classes clearly separate in this embedding, the subsequent classifier will have an easy task performing classification at a high accuracy. We plot this embedding for each layer in the TTN in Fig.~\ref{fig:tsne}. The samples incline to gather in different regions to higher and higher layers, and are finally separated into two curves (1-Dimensional manifold) after the top layer.
	
	% \begin{figure}[htbp]
	% 	\centering
	% 	\label{fig:mnist}
	% 	\includegraphics[width=1.3in,height=1in]{N1000_generalization.eps}
	
	% 	\caption{Training and test accuracy as the function of the bond dimensions on the MNIST dataset. The virtual bond dimensions are set equal to input bond dimensions. The number of training samples is 1000 for each pair of classes.}
	% 	\label{fig:mnist}
	% \end{figure}
	
	\begin{table}
		\centering
		\begin{spacing}{1.2}
			% table caption is above the table
			\caption{10-class classification accuracy(\%) on MNIST}
			\label{tab:1}       % Give a unique label
			% For LaTeX tables use
			\begin{tabular}{llllllllllll}
				\hline\noalign{\smallskip}
				model                  &  0     &  1     &  2     &  3     &  4     &  5     &  6     &  7     &  8     &  9     &  10-class \\
				\noalign{\smallskip}\hline\noalign{\smallskip}
				Training &  99    &  99    &  99    &  97    &  98    &  98    &  99    &  98    &  97    &  94       &  98         \\
				Testing &  99    &  98    &  95    &  94    &  91    &  96    &  97    &  93    &  93    &  92       &  95       \\
				Input bond             &  9     &  10     &  9     &  10     &  10     &  9     &  9     &  9     &  9     &  8        &  /        \\
				Virtual bond           &  9     &  10     &  9     &  10     &  10     &  9     &  9     &  9     &  9     &  8        &  /        \\
				\noalign{\smallskip}\hline
			\end{tabular}
		\end{spacing}
	\end{table}

	% \begin{table}
	% \centering
	% \begin{spacing}{1.2}
	% % table caption is above the table
	% \caption{Baseline performance on MNIST \& CIFAR-10}
	% \label{tab:1}       % Give a unique label
	% % For LaTeX tables use
	% \begin{tabular}{lll}
	% \hline\noalign{\smallskip}
	% model       &Benchmark      &  Test accuracy \\
	% \noalign{\smallskip}\hline\noalign{\smallskip}
	% SVM+GK &  MNIST    &  98.6 \\
	% 1-layer NN &  MNIST    &  97 \\
	% KNN+Euclidean  &  MNIST   & 95   \\
	% Boosted Stump   &  MNIST     &  92.3  \\
	% CNN (LeNet4)   &  MNIST     &  98.9  \\
	% Committee of 35 CNN   &  MNIST     &  99.77  \\
	% CNN   &  CIFAR-10     &  82  \\
	% CNN+augmentation   &  CIFAR-10     &  89  \\
	% Bayesian   &  CIFAR-10     &  85  \\
	% \noalign{\smallskip}\hline
	% \end{tabular}
	% \end{spacing}
	% \end{table}
	
	Generalization denotes the ability to perform prediction on new, unseen cases. To test the generalization of TTN, we use the MNIST dataset of handwritten digits. The training set consists of 60,000 ($28\times 28$) gray-scale images, with 10,000 testing examples. For the simplicity of coding, we rescaled them to ($16\times 16$) images so that the TTN can be conveniently built with four layers.
	
	Fig.~\ref{fig:cifar} (b) shows the representation power and generalization with different numbers of training samples. We find an apparent rise of training accuracy, i.e., the representation power, with larger dimensions of the input and virtual bonds. For the generalization power, it can be exhibited by the testing accuracy. The testing accuracy does not necessarily increase with $\chi$ due to the over-fitting. With 1000 training images for example, the TTN of a higher $\chi$ can capture the training samples more accurately (i.e., higher training accuracy), but the testing accuracy decreases for $\chi > 4$. The over-fitting can be suppressed by increasing the number of training samples to, e.g., 4000 samples. Our results mirror the theoretical principles of statistic learning.	
	
	% For the classification of $D$ classes, we define $| \psi_p \rangle =  \hat{\Psi} |p \rangle$ with $|p \rangle$ the vector corresponding to the $p$-th label. We use the following strategy to obtain $D$ TTN states for the classification of $D$ classes. We first label the training samples as ``yes'' or ``no'' for each class. Then we train the TTN $\hat{\Psi}$ as a binary classifier. $\{| \psi_p \rangle\}$ are obtained as $| \psi_p \rangle =  \hat{\Psi} |\text{yes} \rangle$ ($p = 1 \sim D$). We keep $| \psi_p \rangle$ normalized in the algorithm.
	
	Table~\ref{tab:1} shows the accuracies of the 10 one-against-all classifications and the final 10-class classification. The bond dimensions are chosen as the minimum to reach the accuracy around $95\%$ in the one-against-all classifications. Note some baseline results could be found at the official websites \footnote{The official website of MNIST is available at \href{http://yann.lecun.com/exdb/mnist/}{http://yann.lecun.com/exdb/mnist/}}. For the 10-class classification, the $\{\hat{\Psi}^{[p]}\}$ are calculated from the one-against-all classifiers of the bond dimensions presented in Table \ref{tab:1}.
	
	%\subsection*{Encoding images to states: fidelity and entanglement}
	
\begin{figure}[htbp]
		\centering
		\subfigure{
			\label{fig:Fed_Ent_result:a}
			\includegraphics[width=1.7in,height=1.5in]{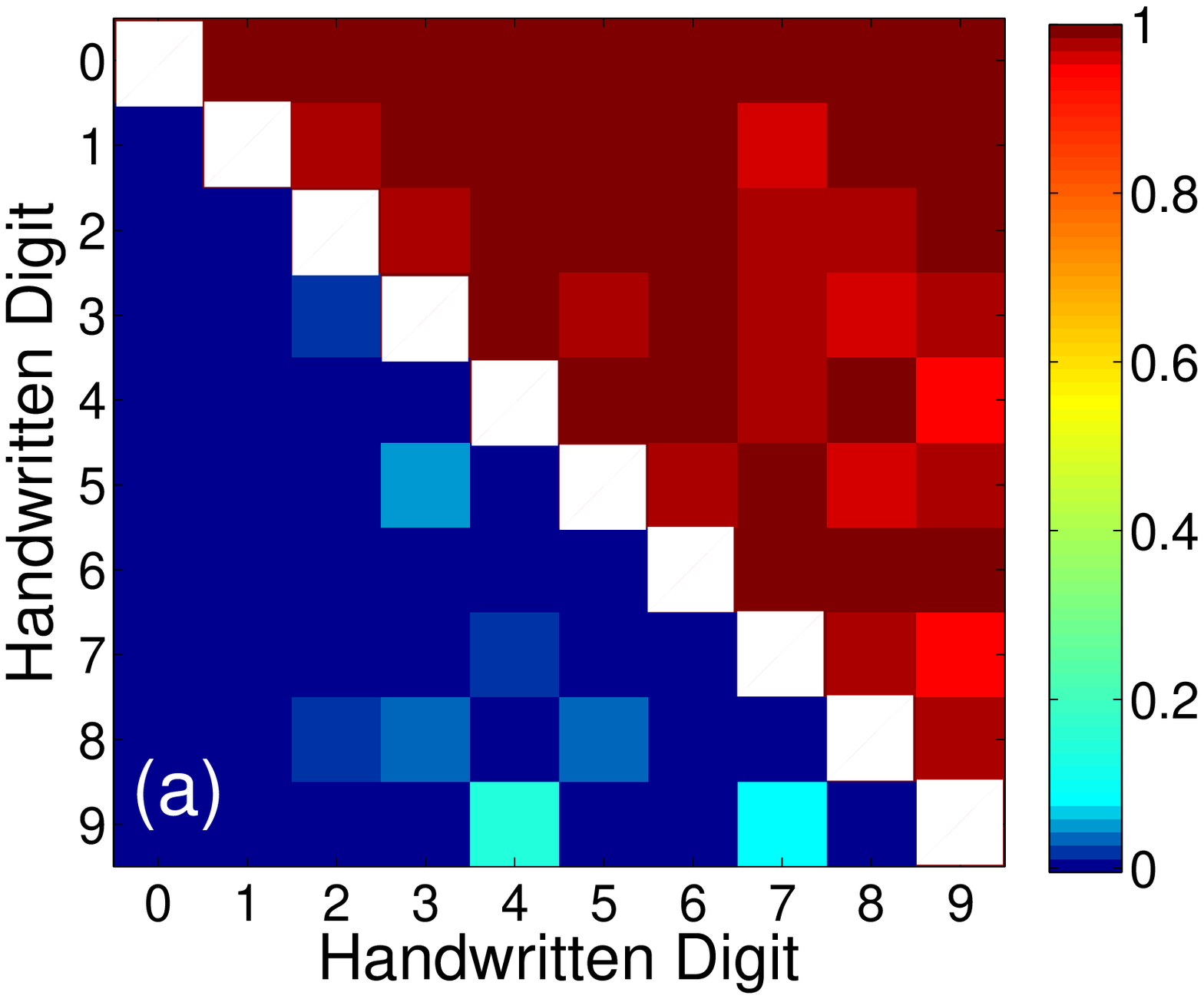}}
		%\hspace{1in}
		\subfigure{
			\label{fig:Fed_Ent_result:b}
			\includegraphics[width=1.5in,height=1.5in]{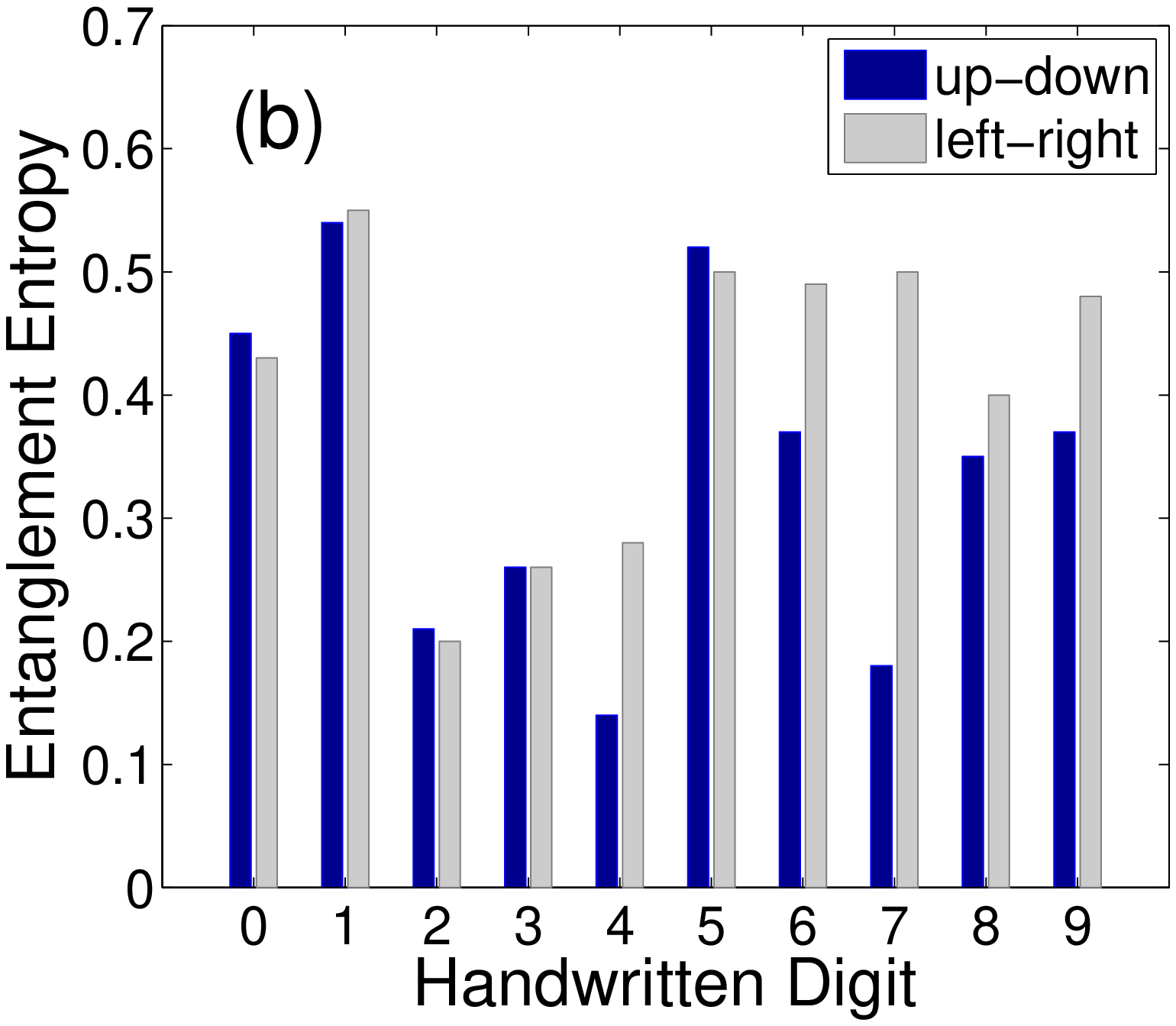}}
		%\hspace{1in}
		\caption{(a) Lower triangular part (colored by blue) shows the fidelity $\mathcal{F}_{p'p}$ between each two classes of handwritten digits. The values range from about $O(10^{-5})$ to $O(10^{-1})$. The upper triangular part (colored by red) shows the binary testing accuracy between each two classes of handwritten digits. (b) Entanglement entropy corresponding to each handwritten digit (input bond = virtual bond = 3).}
		\label{fig:Fed_Ent_result}
	\end{figure}

	\emph{Fidelity and entanglement.}--- The fidelity between two states is defined as $\mathcal{F}_{pp'} = |\langle \psi_{p} | \psi_p' \rangle| $. It measures the distance between the two quantum states in the Hilbert space. Fig.~\ref{fig:Fed_Ent_result:a} shows the fidelity between each two $| \psi_p \rangle$'s trained from the MNIST dataset. One can see that $\mathcal{F}_{pp'}$ remains quite small in most cases. This means that $\{| \psi_p \rangle\}$ are almost orthonormal.
	
	The binary testing accuracy of each two classes is also given in Fig. \ref{fig:Fed_Ent_result:a}. We observe that with larger fidelity, the accuracy is relatively lower, meaning it is more difficult to classify. For example, a large fidelity appears as $\mathcal{F}_{4,9} \simeq 0.14$ with a relatively low accuracy $\mathcal{A}_{4, 9} \simeq 0.95$. We speculate that this is closely related to the way how the data instances are fed and processed in the TTN. The TTN essentially gives a \textit{real-space renormalization group flow}: the input vectors are arranged and renormalized layer by layer according to their spatial locations in the image; each tensor renormalizes four neighboring vectors into one vector. Therefore, two image classes with similar shapes will result in a larger fidelity and higher difficulty to classify. Supplemental to the real-space one, the classification in the frequency space \cite{LZLR18MPLML} might be helpful to further improve the accuracy.
	
	% Fidelity can be potentially applied to building a network, where the nodes are classes of images and the weights of the connections are given by the $\mathcal{F}_{p'p}$. This might provide a mathematical model on how different classes of images are associated to each other.
	
	Another important concept is (bipartite) entanglement~\citep{BD00EntRev}. Entanglement is usually given by a normalized positive-defined vector called entanglement spectrum (denoted as $\Lambda$). In our case, it is the singular value spectrum of the state $| \psi_p \rangle$. With the TTN, the entanglement spectrum is simply the singular values of the matrix $M = T^{[K,1]}|p\rangle$ with $T^{[K,1]}$ as the top tensor. This is because all the other tensors in the TTN are unitary. The strength of entanglement is measured by the entanglement entropy $S = -\sum_{a} \Lambda_a^{2} \ln \Lambda_a^2$. Fig.~\ref{fig:Fed_Ent_result:b} shows the entanglement entropy of $\{| \psi_p \rangle\}$ trained with the MNIST dataset. We compute two kinds of entanglement entropy by cutting the images in the middle along the x and y directions as shown in Fig.~\ref{fig:TTN}. The results were marked by up-down and left-right in Fig.~\ref{fig:Fed_Ent_result:b}. The first one denotes the entanglement between the upper partsof the images with the lower part. The latter denotes the entanglement between the left and the right part.  Note that $M$ has four indexes, of which each represents the effective space renormalized from one quarter of the vectorized image. Thus, the bipartition of the entanglement determines how the four indexes of $M$ are grouped into two bigger indexes before calculating the SVD.
	
	In quantum physics, the entanglement entropy $S$ explicitly reveals the needed dimensions of the virtual bonds to reach a certain precision. Our results show that $\{| \psi_p \rangle\}$ actually possess small entanglement, meaning that the TTN is an efficiently representation of the classifiers. Furthermore in quantum information, $S$ measures the amount of information of one subsystem that can be gained by measuring the other subsystem. An important analog is between knowing a part of the image and measuring the subsystem of the quantum state. Thus, an implication is that the entanglement entropy measures how much information of one part of the image we can gain by knowing the rest part of the image.
	
	%\section*{Discussion}
	\emph{Conclusion and prospective.}--- Focusing on hierarchical 2D TTNs, we propose a MERA-inspired algorithm for machine learning where the tensors are kept to be unitary.  From the numerical experiments with the proposed algorithm, we conclude with the following observations. (1) The upper limit of representation power (learnability) of a TTN depends on the input bond dimensions, and the virtual bond dimensions determine how well the TTN reaches this limitation. (2) A hierarchical TTN exhibits a similar increase level of abstraction as a deep convolutional neural network or a deep belief network. (3) Based on our proposal, We demonstrate the natural connections between classical images and quantum states, which implies that quantum properties (fidelity and entanglement) can be employed to characterize the classical data and the computational tasks.
	
	Moreover, our work contributes to the implementation of machine learning by quantum computations and simulations. Firstly, it has been proposed to implement unitary TTN models as quantum circuits on the quantum hardware~\citep{Huggins2018Towards}. Secondly, by encoding image classes into TTN states, it is possible to realize/certificate machine learning by, e.g., quantum state tomography techniques~\citep{ZHXH+17MPSexp}. 
	
	\emph{Acknowledgments}.--- SJR is grateful to Ivan Glasser and Nicola Pancotti for stimulating discussions. DL was supported by the National Natural Science Key Foundation of China (61433015), the Science \& Technology Development Fund of Tianjin Education Commission for Higher Education (2018KJ217), the China Scholarship Council (201609345008). SJR, PW, and ML acknowledge support from the Spanish Ministry of Economy and Competitiveness (Severo Ochoa Programme for Centres of Excellence in R\&D SEV-2015-0522), Fundaci\'o Privada Cellex, and Generalitat de Catalunya CERCA Programme. SJR and ML were further supported by ERC AdG OSYRIS (ERC-2013-AdG Grant No. 339106), the Spanish MINECO grants FOQUS (FIS2013-46768-P), FISICATEAMO (FIS2016-79508-P), Catalan AGAUR SGR 874, EU FETPRO QUIC, and EQuaM (FP7/2007-2013 Grant No. 323714). S.J.R., C.P., and G.S. are supported by NSFC Grant No. 11834014. S.J.R. acknowledges Fundaci\'o Catalunya - La Pedrera $\cdot$ Ignacio Cirac Program Chair, Beijing Natural Science Foundation (1192005 and Z180013), and Foundation of Beijing Education Committees under Grants No. KZ201810028043. Parts of this work were carried out while PW was employed at ICFO and he acknowledges financial support from the ERC (Consolidator Grant QITBOX) and QIBEQI FIS2016-80773-P), and a hardware donation by Nvidia Corporation. GS and CP were supported by the MOST of China (Grant No. 2018FYA0305800), the Strategic Priority Research Program of the Chinese Academy of Sciences (Grant No. XDB28000000, XDB07010100), the NSFC Grant No. 11834014. CP appreciates ICFO (Spain) for the hospitality during her visit and is grateful to financial support from UCAS and ICFO. This research was supported by Perimeter Institute for Theoretical Physics. Research at Perimeter Institute is supported by the Government of Canada through Industry Canada and by the Province of Ontario through the Ministry of Economic Development and Innovation.
	
	% \bibliography{bibliography}
	%merlin.mbs apsrev4-1.bst 2010-07-25 4.21a (PWD, AO, DPC) hacked
%Control: key (0)
%Control: author (0) dotless jnrlst
%Control: editor formatted (1) identically to author
%Control: production of article title (0) allowed
%Control: page (1) range
%Control: year (0) verbatim
%Control: production of eprint (0) enabled
%

\end{document}